\documentclass[letterpaper]{article} 
\usepackage{aaai24}  
\usepackage{times}  
\usepackage{helvet}  
\usepackage{courier}  
\usepackage[hyphens]{url}  
\usepackage{graphicx} 
\usepackage{natbib}  
\usepackage{caption} 
\usepackage{algorithm}
\usepackage{algorithmic}
\usepackage{amsmath}
\usepackage{newfloat}
\usepackage{listings}
\DeclareCaptionStyle{ruled}{labelfont=normalfont,labelsep=colon,strut=off} 
\floatstyle{ruled}
\newfloat{listing}{tb}{lst}{}
\floatname{listing}{Listing}
\usepackage{subfigure} 
\usepackage{array,booktabs,makecell,tabularx}
\usepackage{enumitem}
\usepackage{footmisc}

\title{SIG: Speaker Identification in Literature via Prompt-Based Generation}
\author {
    Zhenlin Su\thanks{Primary work done in WeChat AI.}\textsuperscript{\rm 1},  
    Liyan Xu\thanks{Co-corresponding authors.}\textsuperscript{\rm 2}, 
    Jin Xu\footnotemark[2]\textsuperscript{\rm 1,3}, 
    Jiangnan Li\textsuperscript{\rm 4}, 
    Mingdu Huangfu\textsuperscript{\rm 1}
}
\affiliations {
    \textsuperscript{\rm 1}School of Future Technology, South China University of Technology\\
    \textsuperscript{\rm 2}WeChat AI, Tencent \\
    \textsuperscript{\rm 3}Pazhou Lab, Guangzhou\\
    \textsuperscript{\rm 4}Institute of Information Engineering, Chinese Academy of Sciences\\
    zhenlinsu75@gmail.com,\, liyanlxu@tencent.com,\, jinxu@scut.edu.cn
}

\begin{document}

\maketitle

\begin{abstract}

Identifying speakers of quotations in narratives is an important task in literary analysis, with challenging scenarios including the out-of-domain inference for unseen speakers, and non-explicit cases where there are no speaker mentions in surrounding context. 
In this work, we propose a simple and effective approach SIG, a generation-based method that verbalizes the task and quotation input based on designed prompt templates, which also enables easy integration of other auxiliary tasks that further bolster the speaker identification performance. The prediction can either come from direct generation by the model, or be determined by the highest generation probability of each speaker candidate. Based on our approach design, SIG supports out-of-domain evaluation, and achieves open-world classification paradigm that is able to accept any forms of candidate input. We perform both cross-domain evaluation and in-domain evaluation on PDNC, the largest dataset of this task, where empirical results suggest that SIG outperforms previous baselines of complicated designs, as well as the zero-shot ChatGPT, especially excelling at those hard non-explicit scenarios by up to 17\% improvement. Additional experiments on another dataset WP further corroborate the efficacy of SIG.


\end{abstract}

\section{Introduction}

Speaker identification in literary text aims at identifying the speaker of quotation in narrative genres such as fictions or novels \citep{DBLP:conf/aaai/ElsonM10}, serving as an important step for downstream applications such as novel-to-script conversion \citep{DBLP:conf/taai/SooYS19}. 

\begin{table}[ht]
\centering
\begin{tabular}{l}
\toprule
\textbf{Quote Type:} Explicit \\
\midrule
\textbf{Quotation} (w/ context): \\
\small{Mrs. Elton hardly waited for the affirmative.}\\
\small{ \textbf{"Well, we shall see."} said Mrs Elton.}\\
\small{Emma was almost too much astonished to answer}\\

\textbf{Speaker}: Mrs Elton  \\
\midrule
\midrule
\textbf{Quote Type:} Anaphoric \\
\midrule
\textbf{Quotation} (w/ context): \\
\small{But Mrs. Elton was very much discomposed indeed.}\\
\small{She said:\textbf{"Rather he than I!"},}\\

\textbf{Speaker}: Mrs. Elton \\
\midrule
\midrule
\textbf{Quote Type:} Implicit\\
\midrule
\textbf{Quotation} (w/ context): \\
\small{"Your father will not be easy; why do not you go?"}\\
\small{\textbf{"I am ready, if the others are."}}\\
\small{"Shall I ring the bell?"}\\
\textbf{Speaker}: Emma \\
\bottomrule
\end{tabular}
\caption{Examples of three quote types in PDNC. The quotation in each example is highlighted in bold.}
\label{tab:type}
\end{table}

Table~\ref{tab:type} shows the examples of three types of quotation in PDNC dataset proposed for this task \cite{DBLP:conf/lrec/VishnubhotlaHH22}, based on whether the speaker is explicitly indicated by an adjoining expression (explicit), or appears without an attribution (implicit) , or is indicated by an anaphoric mention (anaphoric). In this paper, the latter two are referred as the ``non-explicit'' case that cannot be solved by simple superficial narrative patterns, calling for a deep global understanding of the context surrounding the quotation, of which the importance is also highlighted by other related tasks, such as character comprehension in narrative stories \cite{yu2022fewshot,sang-etal-2022-tvshowguess}.
It becomes even more challenging for cross-domain inference, where the system neither has prior knowledge about candidate speakers, nor it could identify the absent speaker mentions in context for those implicit cases. Ideally, a practical system should be able to determine the speakers under any circumstances, achieving an open-world speaker identification paradigm.

\begin{figure*}[t]
\centering
\includegraphics[width=\textwidth]{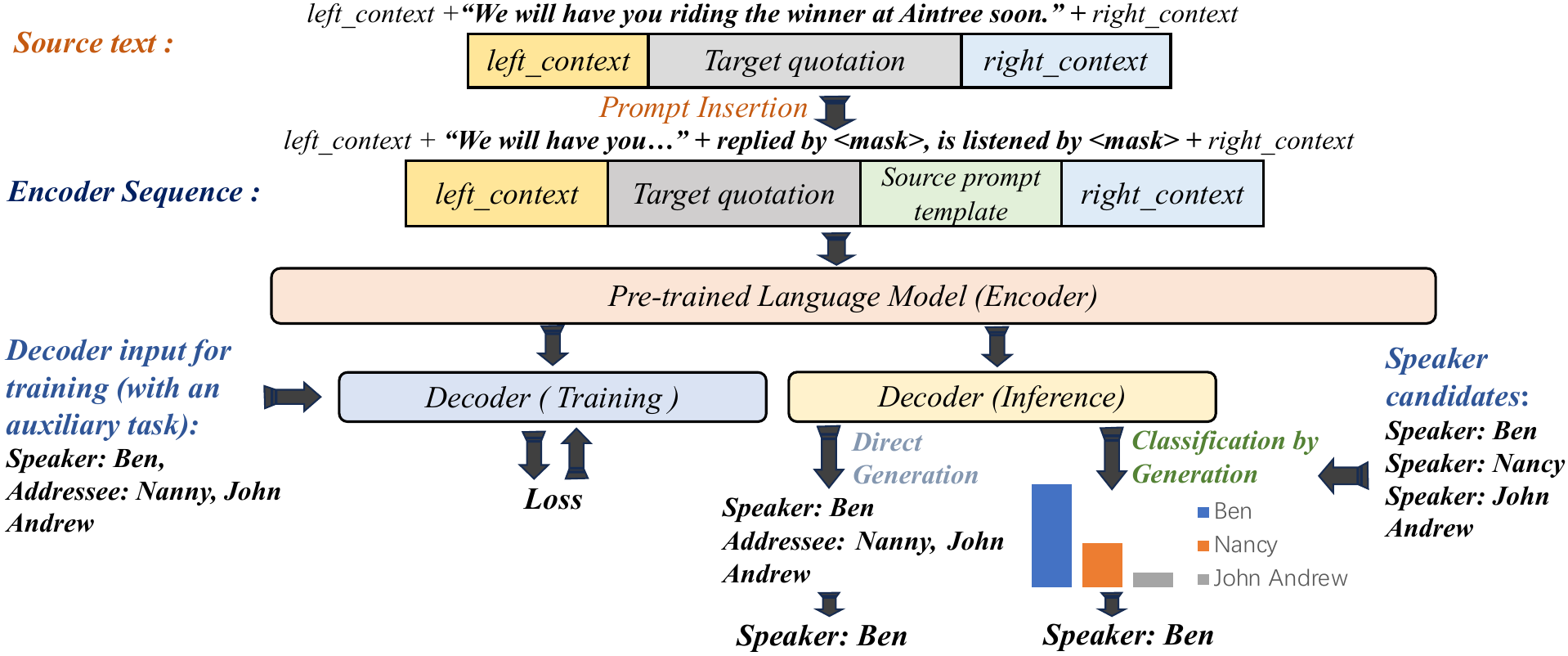}
\caption{Illustration of our proposed approach SIG. Encoder input and decoder output are formatted according to the designed prompt templates described in Section~\ref{ssec:template}. During inference, SIG either generates the speaker directly, or determines the speaker based on the highest generation probability of each candidate.}
\label{figure: framework}
\end{figure*}

Early works on this task involve linguistic-based properties and designs that can be effective for especially the explicit cases \citep{DBLP:conf/aaai/ElsonM10}.
With the remarkable success of pretrained language models (PLMs) such as BERT \citep{DBLP:conf/naacl/DevlinCLT19}, recent studies have shifted their focus towards leveraging PLMs for speaker identification \cite{DBLP:journals/corr/abs-1907-11692}, \citep{DBLP:conf/icassp/PanWYWXM21}.
However, several problems still remain. On the one hand, certain previous works divide the speaker identification task into interrelated subtasks executed with separately trained models, including named entity recognition and coreference resolution (\citep{DBLP:conf/icassp/PanWYWXM21}, \citet{yoder-etal-2021-fanfictionnlp}).
These approaches face inherent challenges due to the inevitable error propagation from subtasks that undermines the final performance \citep{DBLP:conf/naacl/YuZY22}. For instance, the performance of the recent state-of-the-art
coreference resolution model is 85.09 F1 \citep{miculicich-henderson-2022-graph} on the standard CoNLL-12 benchmark \citep{pradhan-etal-2012-conll}, from which the wrong entity clusters will be accumulated to the next step. On the other hand, other previous works regard speaker identification as span extraction to extract the speaker mention spans directly from the input \citep{DBLP:conf/naacl/YuZY22}, which may fail on the implicit cases where such mention spans can not be found in the context surrounding the quotation.
For anaphoric cases, these approaches sometimes are only able to capture pronouns without localizing the explicit speaker identity. Additionally, when dealing with implicit instances, they will always produce incorrect speaker spans, due to the absence of a correct answer to extract.

Recently, prompt-based methods are highlighted by generative Large Language Models (LLMs) such as ChatGPT. To evaluate LLM's capability on this task, we also conduct experiments with zero-shot prompting. Nevertheless, inspired by combining prompt engineering and generative methods, in this work, we propose a new approach for the open-world speaker identification paradigm, which utilizes generation models and converts the task as open-world classification, choosing the speaker according to generation probability, and able to adopt labels in any forms.

Specifically, our approach dubbed SIG (\textbf{S}peaker  \textbf{I}dentification via \textbf{G}eneration), encodes the quotation and its context as the input based on our designed prompt template, then it can either opt to generate the speaker directly, or to evaluate the generation probability for any speaker candidates.
For the latter case, during inference, possible candidate speakers are enumerated, and the speaker with the highest generation probability to fill in the prompt template is selected as the final answer.
In addition, SIG also introduces auxiliary tasks to capture more implicit information related to speaker identification, such as predicting addressees, which is seamlessly integrated into the designed prompt template as well.

Compared to previous methods, SIG can offer notable advantages. Firstly, SIG approaches the speaker identification task in an end-to-end manner, bridging the input and output through the prompt template by generation, which circumvents the inherent drawbacks associated with traditional pipeline approaches comprising multiple sub-tasks. Secondly, it is more flexible to handle a variety of situations, including the implicit cases where the correct answer does not appear in the context, as well as the cross-domain cases where the speaker is not seen by the model during training.

Our empirical experiments are mainly conducted on the Project Dialogism Novel Corpus (PDNC) \citep{DBLP:conf/lrec/VishnubhotlaHH22}, of which about 66\% are the non-explicit cases. The results demonstrate that SIG outperforms the state-of-the-art approaches on PDNC by a significant margin of 4\% averaged accuracy for all quotations, and a substantial 17\% averaged accuracy specifically for non-explicit cases. In addition, ChatGPT presents strong performance; our approach still outperforms zero-shot ChatGPT (GPT-3.5-turbo) by 9\% for all quotations and 2\% for non-explicit quotations. 
These results indicate that our method achieves state-of-the-art performance on the cross-domain speaker identification task on PDNC, especially for complex cases.
In addition, SIG is also evaluated on WP \cite{chen2019chinese}, a speaker identification dataset stemmed from a Chinese novel, and shown surpassing previous baselines by 5.2\% on the test set.
It is worth noting that although this work does not use LLMs for training due to the computational resource constraints, our proposed approach could adopt LLMs in future for further enhancement.

Overall, our contributions in this work are four-fold:
\begin{itemize}[noitemsep,nolistsep,leftmargin=*]
    \item We propose SIG, an approach for speaker identification via prompt-based generation, achieving open-world classification on speakers, to be a new paradigm for this task.
    \item Experiments suggest that SIG outperforms existing methods as well as zero-shot ChatGPT on PDNC, demonstrating superior performance on cross-domain evaluation and the harder non-explicit cases.
    \item SIG is also shown to outperform baselines on WP for the in-domain evaluation setting, showing robust performance of our proposed approach.
\end{itemize}

\section{Background and Related Work}


\paragraph{Speaker Identification}
Previous works have proposed several datasets for evaluation. \citet{DBLP:conf/aaai/ElsonM10} introduces the CQSA corpus, which contains quotations from 4 novels and 7 short stories that are annotated. \citet{DBLP:conf/lrec/BammanLM20} released LitBank corpus, an annotated dataset of 100 English fictions, aimed at supporting various tasks including entity recognition, coreference resolution, and speaker attribution.
The recently proposed Project Dialogism Novel Corpus (PDNC) \citep{DBLP:conf/lrec/VishnubhotlaHH22} comprises 22 comprehensive works of fiction, encompassing 12,773 explicit quotations and 23,205 non-explicit quotations, being the largest dataset for training and evaluation.

\paragraph{Previous Approaches}
\citet{DBLP:conf/aaai/ElsonM10} first introduces quote attribution in literary narratives by assigning speaker tags to quoted speech. They utilize rule-based and statistical learning techniques to identify candidate characters, determine their genders, and attribute each quote to the most likely speaker.
\citet{DBLP:conf/eacl/JurafskyCMF17}  proposes a deterministic, two-step methodology for quotation attribution. A sequence of progressively intricate filters are utilized to initially associate each quotation with a mention, and then it proceeds to link the mention to a character entity.

Moving to the era of PLMs, \citet{DBLP:conf/interspeech/ChenLL21} proposes a candidate scoring network based on BERT along with a revision algorithm to identify the speaker. \citet{yu-etal-2022-end} resolves speaker identification as a span extraction task, and applies the extractive Question-Answering paradigm to address it.
Pipeline approaches are also mainstream \cite{DBLP:conf/lrec/BammanLM20,xu-choi-2022-modeling}, utilizing BERT to independently train base models for modules such as coreference resolution \cite{xu-choi-2020-revealing}.
\citet{DBLP:conf/lrec/VishnubhotlaHH22} enhances this approach by restricting the set of candidates to resolved mention spans from the coreference resolution step for direct quotation-to-entity linking, resulting in state-of-the-art performance on PDNC. These studies highlight the significance of PLMs for speaker identification, while also still leave room for improvement due to several disadvantages as discussed above.

\paragraph{Template-based Approach}
\citet{DBLP:conf/nips/BrownMRSKDNSSAA20} is among the first works to utilize prompts for solving text classification tasks. \citet{DBLP:journals/corr/abs-2001-07676} further employs a template-based approach to transform text classification into a cloze-style problem, by training the model to fill in the provided slots. \citet{DBLP:conf/acl/CuiWLYZ21} has formulated templates for both input and output, selected the optimal choice by the option of the most likely generation, addressing the few-shot Named Entity Recognition (NER). The generative paradigm has also shown successful in other NLP tasks \cite{7475902,lu-etal-2021-text2event}. In this work, we adopt the generative paradigm for open-world classification that has been advocated in other areas \cite{xu-etal-2023-towards}. The success of these approaches in other areas demonstrates the feasibility of incorporating specific prompt templates for task verbalization aligned with the pretraining stage.

\section{Methodology}
\label{sec:approach}

\subsection{Task Definition}

Before delving into the specifics of our proposed approach, the basics of the speaker identification task are clarified as follows. Within the context of narrative corpus, sentences can be categorized into two distinct types: those containing direct speech by characters (referred to as quotations), and those comprising narrative descriptions. Our task revolves the extraction of pertinent information from the contextual surroundings of quotations that is subsequently employed to determine the most likely speaker. The names and aliases of potential speakers can be pre-collected. Our approach targets to identify the speaker for a given quotation accompanied by its surrounding context.

\subsection{Approach Introduction}
\label{ssec:template}

Figure~\ref{figure: framework} illustrates our proposed approach SIG. It takes the quotation along with its left and right context as template-input, generating the speaker directly, or a score for each candidate speaker. To initiate the process, the source prompt template is inserted with a placeholder \textless mask\textgreater \;positioned after the quotation.
Guided by the target prompt template, the model encodes the designed input and starts the auto-regressive mechanism for generation to fill the speaker candidate in the target template.

As shown in Figure \ref{figure: framework}, SIG inserts the prompt containing the placeholder after the quotation, and defines the task flexibly verbalized by natural language, which minimizes the gap between the pretraining and finetuning for models such as BART \cite{lewis-etal-2020-bart}, and helps to better leverage the internal knowledge of PLMs.

\paragraph{Prompt Template Design}
Different prompt templates are examined as shown in Table~\ref{tab:template}. Concretely, both the source (input) and target (output) can be augmented with prompts such as ``\textit{replied by:}'' or ``\textit{speakers:}''.
For the source template, it is followed by \textless mask\textgreater , processed by the encoder; while for the target, the decoder is expected to fill with the correct speaker. Additionally, we also experiment with the naive version without any templates, where the decoder generates the speaker directly based on the encoded quotation and context, without using the placeholder \textless mask\textgreater.
These prompts verbalize the task natually and instruct the model to pay more attention towards the quotation.

\paragraph{Prompt Template with Auxiliary Task}
One of the advantages for prompt-based task verbalization is that we can easily integrate other tasks together by adding more instructions to the template.
In this work, we also investigate incorporating the auxiliary task for speaker identification. Here, we train the model to simultaneously recognize both the speaker and the addressees of a quotation, learning to disentangle the multi-party interactions in a dialogue, which may potentially bolster the quotation attribution itself. Moreover, given that both the addressees and the speaker are typically represented by personal names, predicting the addressees strengthens the model's attention towards the pertinent person names in the context.


In our approach to integrate with the auxiliary addressee prediction, we simply extend the prompt by adding ``\textit{is listened by} \textless mask\textgreater'' in the source template, and ``\textit{Addressee:}'' in the target template, as depicted in Figure~\ref{figure: framework}.
For the inference, the model will still produce the speaker first, and the predicted addressees could be discarded.

Table~\ref{tab:aux} also shows results with other auxiliary tasks, e.g. gender identification. SIG adopts addressee prediction as it leads to the best result empirically.

\subsection{Inference}
\label{ssec:inference}

\paragraph{Direct Generation}
As SIG operates upon the generative model, it is straightforward to generate the speaker directly.
We take the quotation and its context as the input, formatted by the prompt template, and model generates the speaker of the quotation without decoding constraints.
The predicted speaker is then parsed from the generated output.
For evaluation, it is judged whether it belongs to one of the speaker candidates or aliases from gold labels. 
However, this process could potentially introduce the following disadvantages. First, the generated output sometimes does not match the corresponding speaker accurately. For example, if the string generated by the model is \textit{Beaver}, it is difficult to resolve whether it refers to \textit{Mrs. Beaver} or \textit{Mr. Beaver}. Second, it is harder to control the generation, especially when the model also adopts other auxiliary tasks. 

\paragraph{Classification by Generation}
To address the drawbacks in direction generation, we propose to guide the generation process by providing the speaker candidates to the decoder.
We begin by listing all candidate speakers for the given quotation, and for each candidate, we obtain its generation probability according to the trained model. Thus, this paradigm supports any forms of speakers, including new speakers unseen during training, or speakers that do not appear in the surrounding context.

As speaker names can often be of different lengths, in order to cope with this situation, we take the averaged probability of the output, and select the one with the highest probability to be the final prediction. The score for the $i$th candidate can be denoted as follows:


\begin{equation}
\label{eq2}
f(\mathbf{T}_{i}) = \sum_{c=1}^{m} p(t_{c}|t_{1:c-1}, \mathbf{X}) / m,
\end{equation}
where $T_i = (t_1, .., t_m)$ is the target template output with length $m$, $X$ is the source input, and $p$ represents the probability from the decoder at each step.

\subsection{Training}

Given a speaker quotation $\mathbf{q}$, its left context $\mathbf{c}^{l}$ and right context $\mathbf{c}^{r}$, the source prompt template $\mathbf{t_s}$, the input for the model $\mathbf{X}$ is formulated as follows:
\begin{equation}
        \text{[CLS]} \oplus \mathbf{c^l} \oplus \mathbf{q} \oplus \mathbf{t_{s}} \oplus \mathbf{c^r} \oplus \text{[SEP]}, 
\end{equation}

where $\oplus$ represents concatenation (example in Figure~\ref{figure: framework}).
The model encodes the provided input, and the output is conditioned on the hidden state of the input, following the standard encoder-decoder architecture.
The training adopts teacher-forcing, where the model is optimized to minimize the negative log-likelihood of the gold output sequence, denoted as below:

\begin{equation}
    \mathcal{L} = - \sum_{c=1}^{m} \mathrm{log}\; p(t_{c}|t_{1:c-1},\mathbf{X}) 
\end{equation}

The training is the same for either direction generation or classification by generation. Thus, the trained model is flexible to choose the inference according to specific scenarios.

\begin{figure}[th]
\centering
\includegraphics[width=0.65\columnwidth]{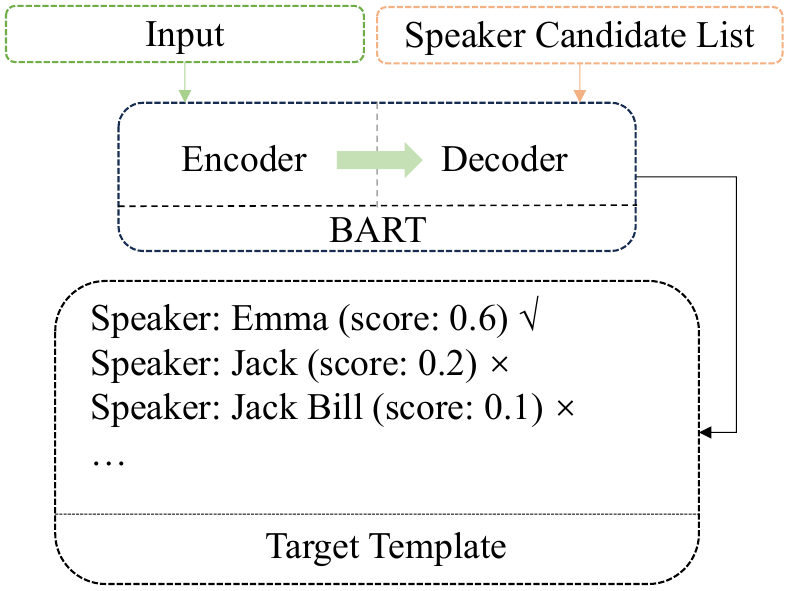}
\caption{Classification by generation: each speaker candidate is fed to the decoder, and its generation probability is obtained; the highest option is selected as the final prediction.}
\label{figure: score}
\end{figure}

\section{Cross-Domain Experiments}
\label{sec:experiments}

\subsection{Dataset}
\label{ssec: dataset}
Our main experiments are conducted on Project Dialogism Novel Corpus (PDNC) \citep{DBLP:conf/lrec/VishnubhotlaHH22}, a recently introduced dataset specifically designed for analysis on English literary text. PDNC stands out as the largest dataset for this task, encompassing a diverse range of genres such as science fiction, literary fiction, children's literature and detective fiction, each contributing to a stylistically varied speech style.

To elaborate, PDNC consists of a compilation of 22 extensive novels, comprising a total of 35,978 identified and annotated quotations that enable training. These annotations capture crucial attributes including the speaker, addressees, and quotation type. Quotations within PDNC are categorized into three distinct types as introduced in Table~\ref{tab:type}: explicit, implicit, anaphoric.


\subsection{Evaluation Protocol}
\label{ssec:protocol}

As PDNC consists of multiple novels, it enables cross-domain evaluation, such that the model is evaluated on the test set that has no overlapped novels from the training. It is a more practice scenario, which is also adopted by PDNC authors \cite{DBLP:conf/lrec/VishnubhotlaHH22,DBLP:conf/acl/VishnubhotlaRHH23}, as the trained model should be able to recognize speakers upon any literary, rather than only for those limited novels seen during training.

Following the evaluation outlined by previous works, our results are reported as averaged accuracy over five experiments. For each experiment, four novels are randomly selected as the test set, while the remaining novels are used for training the model. The process ensures that no novels are select twice in the test set.

In addition, previous works also exclude instances from training and evaluation where speakers have fewer than 10 annotated quotations. This step was taken to mitigate the potential impact of minor characters, which often fall under the long-tail distribution, thereby ensuring a more focused and insightful analysis. Our experiments keep the same setting, to be comparable with previous approaches.



\subsection{Experimental Approaches}
\label{ssec:cross-approach}

\paragraph{BookNLP}
BookNLP is a NLP tool tailored for English literary text. Its processing pipeline encompasses various tasks, including but not limited to named entity recognition, conference resolution, and speaker attribution. During the speaker attribution phase, BookNLP initiates the process by employing the entity recognition and conference resolution models to identify person clusters within a specified window surrounding the quotation. A BERT model incorporating contextual, positional, and gender cues then assigns scores to mention spans located within this window. The mention span with the highest score is then selected as the attributed speaker.

To ensure alignment between the extracted mention spans from BookNLP and the labels within the PDNC corpus, we suitably relax the matching criteria. Any mention span that corresponds to a mention span within the PDNC is considered correct. Additionally, if an incorrect mention span is clustered to the correct answer, it is still considered a valid choice.

\paragraph{BookNLP+}
In a complementary effort, \citet{DBLP:conf/acl/VishnubhotlaRHH23} enhances BookNLP by constraining the set of candidates to resolved mention spans stemming from the coreference resolution step. This refined approach directly establishes a link between quotations and entities, achieving state-of-the-art results for the cross-domain evaluation on PDNC.

\paragraph{Large Language Models (LLMs)}
We employ ChatGPT (\textit{gpt-3.5-turbo-0613}) from OpenAI for the zero-shot experiments. Certain efforts for prompt engineering are performed to refine the task prompts. Especially, Chain-of-Thought (CoT) \cite{cot} is also adopted for all LLM experiments. 
Due to the expense of API usage, we conduct only a single evaluation of ChatGPT.

Since LLMs are not trained for this task, we employ a lenient metric for evaluation. A response is considered correct as long as the true speaker's name (or one of its aliases) appears as a substring in the response. Thus, our reported results for ChatGPT can be considered as an upper bound performance under our specific prompt setting.

\paragraph{RoBERTa}
We also adopt RoBERTa \cite{DBLP:journals/corr/abs-1907-11692} as the encoder-only model for the conventional classification paradigm, where a linear layer is stacked on the last layer of hidden state, followed by softmax to classify speakers directly.
We keep the same prompt template as SIG for the source input.
However, since this method can only perform in-domain evaluation, and cannot handle speakers not seen during training, we report the in-domain evaluation results separately from the main cross-domain evaluation.

\paragraph{SIG\textsuperscript{D}}
This method is the \textbf{D}irect generation setting of our proposed approach, described in Section~\ref{ssec:inference}.

\paragraph{SIG}
This method adopts the classification as generation setting of our approach, which is more flexible than SIG\textsuperscript{D}. Both SIG\textsuperscript{D} and SIG employ BART \cite{lewis-etal-2020-bart} as the sequence generation PLM.

\subsection{Results and Discussions}

\begin{table}[htbp!]

    \centering
    \begin{tabular}{l|c c c c c c c}
    \toprule
         & \bf Non-Explicit & \bf Total\\ 
       \midrule
        BookNLP  & 0.46/0.39 & 0.68/0.66 & \\
        BookNLP+ & $0.53^{*}$/-  & $0.68^{*}$/- \\
        \midrule
        ChatGPT  & 0.70/0.71 & 0.71/0.71\\
        \midrule
        SIG\textsuperscript{D} & 0.56/0.51 & 0.57/0.54\\
        SIG & \textbf{0.70}/\textbf{0.73} & \textbf{0.72}/\textbf{0.78} \\
        \bottomrule
    \end{tabular}
    \caption{Accuracy for the cross-domain evaluation on PDNC, broken down by quotation types. For BookNLP+, we take the reported accuracy from \citet{DBLP:conf/acl/VishnubhotlaRHH23}. For each other approach, we report both the mean and median accuracy of the five repeated experiments described in Section~\ref{ssec:protocol}. Our proposed approach SIG is shown to obtain the best accuracy.}
    \label{tab:cross-fiction-2}
\end{table}

\begin{table*}[thbp!]
    \centering
    \begin{tabular}{l |c|c}

\toprule
       \textbf{Source Template} & \textbf{Target Template } & \textbf{Accuracy}\\
       \midrule  
       
       No Source template &No target template  & 56.34\\
       ``Quotation + replied by: \textless mask\textgreater" &No target template  & 58.12\\
       ``Quotation + replied by: \textless mask\textgreater" &  replied by:
   \textless Candidate\_speaker\textgreater  &66.32 \\
       ``\textbf{Quotation + replied by: \textless mask\textgreater}'' &\textbf{Speaker: \textless Candidate\_speaker\textgreater}  & \textbf{68.43}\\
       ``Quotation + Speaker: \textless mask\textgreater" & replied by: \textless Candidate\_speaker\textgreater  & 58.14\\
       ``Quotation + Speaker: \textless mask\textgreater" &Speaker: \textless Candidate\_speaker\textgreater & 64.44\\
       \bottomrule
  
    \end{tabular}
    \caption{Results on PDNC using different prompt templates with BART, described in Section~\ref{ssec:template}. Though these prompts are quite simple, evaluation suggests that they indeed have quite an impact to the final performance. SIG adopts the prompt template of the best performance highlighted by bold.}
    \label{tab:template}
    
\end{table*}


Table~\ref{tab:cross-fiction-2} shows the cross-domain evaluation results with approaches described in Section~\ref{ssec:cross-approach}.
Several observations could be made as follows.

First, Table~\ref{tab:cross-fiction-2} suggests that SIG surpasses all baselines, including the zero-shot ChatGPT, for both non-explicit quotations or for all quotation types, achieving the new state-of-the-art for the cross-domain setting.  Notably, our proposed method demonstrates an improvement of 4\% for all types, and especially a significant 17\% for the non-explicit quotations.
The superior results underscore SIG's robust generalization capability to identify speakers in unseen domains and novels, as well as recognizing the speakers for non-explicit quotations beyond the superficial cues. 

Second, ChatGPT obtains impressive results, despite it has not undergone any task-specific training for PDNC. Particularly, Table~\ref{tab:cross-fiction-2} shows that by simply performing zero-shot inference along with appropriate prompt engineering, ChatGPT outperforms both BookNLP and BookNLP+ that are of sophisticated approach design by good margins. It indicates that LLM-based methods are future-proofing, such that they either resolve tasks directly through their powerful understanding and reasoning abilities, or can be used for training that serves as a direct substitution of conventional generative models such as BART. In this work, we do not adopt training with open-source LLMs due to constraints of our computational resources.


\subsection{Ablation Studies}
\paragraph{Prompt Templates}
Table~\ref{tab:template} shows evaluation results with various prompt templates for speaker identification without adding auxiliary tasks.
Notably, by judiciously selecting an appropriate pair of prompt templates, the performance receives a substantial increase of \textbf{12.1\%} compared to not using any prompts at all. In light of these outcomes, SIG adopts the best-performing template from these subsequent experiments. The results further corroborates the importance to align the finetuning task to the model's pretraining stage to better induce its internal knowledge.

\paragraph{Auxiliary Tasks}
The performance of speaker identification accompanied by different auxiliary tasks is depicted in Table~\ref{tab:aux}. As observed, training with either addressee identification (71.59\%) or gender identification (69.51\%) yields enhanced results compared to training without any auxiliary tasks (68.43\%). Conversely, when the model is trained with fiction classification to identify which novels the quotation comes from, there is a noticeable decline by more than 5\%. These findings underscore the notion that not all auxiliary tasks are able to boost the main task of speaker identification. Consequently, a careful selection of appropriate auxiliary tasks is imperative to optimize performance.

\begin{table}[htbp!]
    \centering
    \begin{tabular}{l |c}

       \toprule
       \textbf{Auxiliary Task} &\textbf{Accuracy}\\
       \midrule  
       
       None & 68.43\\
       Fiction Identification & 63.38\\
       Gender Identification  & 69.51\\
       \textbf{Addressee Identification} & \textbf{71.59}\\
       \bottomrule
  
    \end{tabular}
    \caption{Results on PDNC adopting different auxiliary tasks. SIG employs the addressee prediction as the auxiliary task, as it achieves the best performance.}
    \label{tab:aux}
\end{table}

\section{In-Domain Evaluation}


\subsection{PDNC Experiments}
Distinct from the cross-domain evaluation in Section~\ref{sec:experiments}, in-domain evaluation provides the same set of novels for both training and evaluation. Thus, speaker candidates remain the same, and the trained model gains prior knowledge regarding those speakers after finetuning.

Though the in-domain evaluation is a less practical setting, as the model may not be able to recognize unseen speakers, it can be regarded as an upper bound of the model capability, since for the same novel, a model with prior speaker knowledge is likely to outperform the model that is trained on other novels.
For this setting, we compare SIG with the conventional classification paradigm, where we utilize RoBERTa, described in Section~\ref{ssec:cross-approach}, to directly classify the attributed speaker for a quotation during evaluation.

Following the setting outlined in \citep{DBLP:conf/lrec/VishnubhotlaHH22}, the training and test set for each novel is organized based on quotation types. Specifically, the training set comprises only explicit quotations, while the remaining quotations (implicit and anaphoric) are assigned to the test set. This is a harder setting than randomly splitting quotations as training or evaluation, as now the model does not see any non-explicit quotations during training, which requires generalization of the model capacity. The overall statistics for cross-domain and in-domain evaluation on PDNC is provided in Table~\ref{tab:stats}.

\begin{table}[htbp!]
    \centering
    \begin{tabular}{l|cc}
       \toprule
       & \textbf{Train} &\textbf{Test}\\
       \midrule  
       Cross-Domain & 28105 (32.8\%) & 5989 (34.7\%) \\
       In-Domain & 11235 (100\%) & 22589 (0\%) \\
       \bottomrule
    \end{tabular}
    \caption{The number of quotations of the training and test set on PDNC for the cross-domain evaluation and in-domain evaluation. The ratio of explicit quotations is shown in parentheses.}
    \label{tab:stats}
\end{table}

\begin{table}[htbp!]
    \centering
    \begin{tabular}{l |c}

       \toprule
        & \textbf{Top 1-5 Accuracy}  \\
       \midrule  
       RoBERTa  & \textbf{69.39} / \textbf{78.56} / \textbf{80.23} / 82.45 / 84.55 \\
       SIG & 64.45 / 75.32 / 78.36 / \textbf{83.16} / \textbf{86.31} \\
       \bottomrule
    \end{tabular}
    \caption{The performance for the in-domain speaker identification. Top 1-5 refers to the correct answer being within the top 1 to top 5 predictions by the model.}
    \label{tab:id}
\end{table}

\begin{figure}[htbp!]
\centering
\subfigure[Output embedding from RoBERTa.]{
\includegraphics[width=0.3\columnwidth]{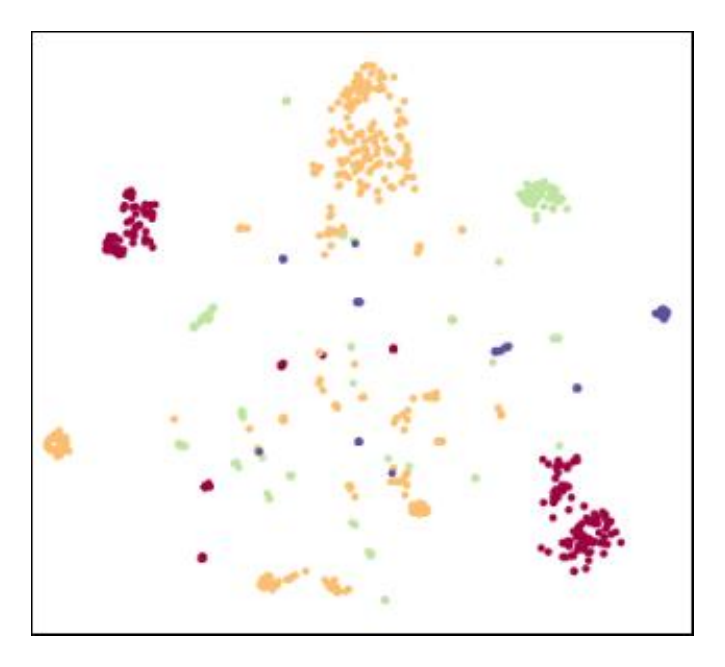} \label{v1}
}
\quad
\subfigure[Speaker embedding from SIG.]{
\includegraphics[width=0.3\columnwidth]{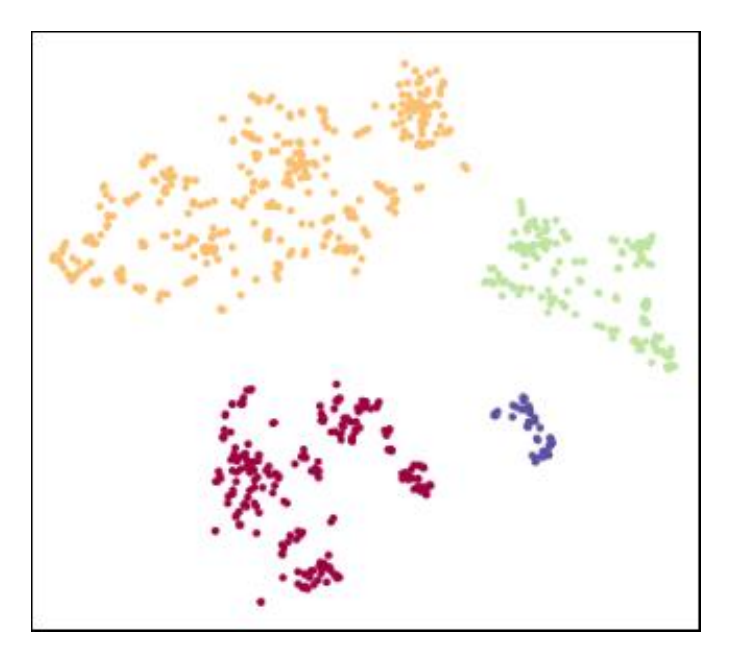} \label{v2} 
}
\caption{t-SNE visualization of the embedding distribution on the test set for PDNC. Output from the same novel is marked by the same color.}
\label{visualization}
\end{figure}

Table~\ref{tab:id} shows the evaluation results by top-k accuracy of RoBERTa and SIG. Furthermore, Figure~\ref{visualization} provides the t-SNE visualization of the output embedding from both approaches. For RoBERTa, it is the averaged embedding of quotation that is used for classification; for SIG, it is the decoder embedding of the speaker names.

Upon analyzing the results in Table~\ref{tab:id} and Figure~\ref{visualization}, we could arrive at the following conclusions:
\begin{itemize}
    \item Table~\ref{tab:id} shows that even compared to RoBERTa that is designed specifically for in-domain speaker identification, SIG lags behind by less than 5\% without any modification, and starts to surpass Roberta consistently when k $\geq $ 4 (by 0.7\% when k = 4, with larger margin when k is larger). Importantly, SIG is also ready to perform cross-domain evaluation that RoBERTa cannot handle at all.
    \item By Figure~\ref{visualization}, it becomes apparent that the output embedding of SIG exhibits clustering based on novels, whereas certain embedding of RoBERTa is notably distant from the clusters. This observation suggests that SIG is adept at capturing the relationships associated with the labels of candidate speakers.
\end{itemize}

\subsection{WP Experiments}
\label{ssec:wp}

WP \cite{chen2019chinese} is a dataset annotated on the Chinese novel \textit{World of Plainness}. The name list was collected manually and contained 125 roles that occurred throughout the novel. \citet{chen2021neural} further extended this dataset by making additional annotations and obtained 2596 quotation instances in total. WP only supports in-domain evaluation as it only comprises one novel.

\paragraph{Baselines}
\begin{itemize}
\item Candidate scoring network (CSN) \cite{chen2021neural}: this method follows a three-step pipeline, nearest mention location(NML), candidate scoring network(CSN) and speaker alternation pattern. Initially, an input instance accompanied by the name list is sent to NML to obtain a set of speaker candidates.
Subsequently, each candidate is sent to CSN to generate its score.
A revision based on SAP is then implemented on the quotation that produces the final decision.
\item End-to-End Speaker Identification (E2E-SI) \cite{yu-etal-2022-end}: for each quotation input, this method extracts speaker spans appeared in surrounding context, similar to the extractive Question-Answering model, where it identifies start and end positions of the predicted speaker spans for the final decision.
\end{itemize}

Since WP does not provide additional information such as addressees, all models are trained for speaker identification only, without other auxiliary tasks.

\paragraph{Results}
Table~\ref{tab:chinese-result} shows the evaluation results on WP. SIG outperforms the two baselines by 3.6\% and 5.2\% respectively. Particularly, SIG\textsuperscript{D} obtains comparable performance as SIG, indicating that even with direct generation without seeing the speaker candidates, the simple generation paradigm with prompt templates could still be superior to the pipeline or encoder-only approaches.

\begin{table}[htbp!]
    \centering
    \begin{tabular}{l|c c c c c c c}
    \toprule
       & Dev & Test \\ 
       \midrule
        CSN & - & $82.50$ \\
        E2E-SI & $78.60$ & $80.90 $ \\
        SIG\textsuperscript{D} & 85.27 & 85.89\\
        SIG & \textbf{85.81} & \textbf{86.15} \\ 
        \bottomrule
    \end{tabular}
    \caption{Accuracy for the in-domain evaluation on WP. Baselines are described in Section~\ref{ssec:wp}, and their results are directly taken from the original papers.}
    \label{tab:chinese-result}
\end{table}

\section{Conclusions}
In this work, we propose SIG, an approach that supports the open-world speaker identification on literary text. SIG adopts the generation-based method, verbalizing the task and quotation input based on designed prompt templates; especially, SIG is flexible integrating other auxiliary tasks by simply extending the prompt. The inference can be either direct generation, or select the speaker candidate with the highest generation probability, enabling inference on different domains whose speakers are not seen by the model during training.
Evaluation on PDNC demonstrates that SIG surpasses all baselines, as well as the zero-shot ChatGPT, particularly excelling in cross-domain non-explicit speaker identification scenarios that demand a profound comprehension of context. Additional in-domain evaluation on PDNC and WP further confirms the efficacy of SIG. 

\section*{Acknowledgments}

This work is supported by National Natural Science Foundation of China (62372187) and Guangdong Provincial Key Laboratory of Human Digital Twin (2022B1212010004).

\bibliography{aaai24}

\end{document}